%% file: acl2023.tex
\pdfoutput=1

\documentclass[11pt]{article}

\usepackage[]{ACL2023}

\usepackage{times}
\usepackage{latexsym}

\usepackage[utf8]{inputenc} 
\usepackage[T1]{fontenc}    
\usepackage{url}            
\usepackage{booktabs}       
\usepackage{amsfonts}       
\usepackage{nicefrac}       
\usepackage{microtype}      
\usepackage{amsmath,amssymb}
\usepackage{epsfig,graphicx,subfigure,caption}
\usepackage{algpseudocode}
\usepackage{multirow,hhline}
\usepackage[ruled,noend]{algorithm2e}
\usepackage{amsmath, bm}
\usepackage{wrapfig}
\usepackage{lipsum}
\usepackage{caption}
\usepackage{footnote}
\usepackage[bottom]{footmisc}
\usepackage[toc,page]{appendix}
\newcommand{\baby}{CCLM\xspace}
\newcommand{\babymmm}{CCLM$_\mathrm{base}^\text{3M}$\xspace}
\newcommand{\babyB}{CCLM$_\mathrm{base}$\xspace}
\newcommand{\babyL}{CCLM$_\mathrm{large}$\xspace}

\usepackage[T1]{fontenc}

\usepackage[utf8]{inputenc}

\usepackage{microtype}

\usepackage{inconsolata}

\title{Cross-View Language Modeling: Towards Unified Cross-Lingual Cross-Modal Pre-training}

\author{
  Yan Zeng\thanks{\, Equal Contribution. Work done at ByteDance.}\;\thanks{\, Correspondence to: zengyan.yanne@bytedance.com} \\ 
  ByteDance \And
  Wangchunshu Zhou$^*$ \\
  ETH Zurich \AND
  Ao Luo$^*$ \\
  Waseda University \And
  Ziming Cheng$^*$ \\
  Shanghai Jiao Tong University \And
  Xinsong Zhang \\
  ByteDance
  }

\begin{document}
\maketitle

\input{sections/0_abstract}
\input{sections/1_introduction}
\input{sections/2_related}

\input{sections/3_method}
\input{sections/4_experiments}

\input{sections/5_conclusion}

\section*{Limitations}
In this paper, we pre-train \baby with moderate multi-modal data, e.g. CC3M, to make a fair comparison with previous work such as M$^3$P and UC$^2$. We leverage large-scale vision language pre-training simply by utilizing the pre-trained weights of X$^2$-VLM which has been pre-trained on billion-scale image-text pairs in English. Collecting more image-text pairs in different languages  will very likely lead to further performance improvements. Moreover, there exists larger public available multi-lingual datasets, such as MultiUN~\citep{ziemski2016united} and OPUS~\citep{tiedemann2012parallel}. Leveraging more multi-lingual datasets for pre-training should also yield a more powerful multi-lingual multi-modal model.


As for social impact, multi-modal pre-trained models can be used in applications that help people with disability in one modality. Our work makes these applications applicable to minority people speaking non-English, and potentially low-resource languages. In sum, our work potentially enables deep learning technology to benefit more people, and is unlikely to have direct negative social impact.

\section*{Acknowledgements}
We would like to thank Hang Li, Jiaze Chen, and Huiyun Yang at ByteDance for insightful comments in technical discussions. We also thank Yaoming Zhu at ByteDance for his generous assistance in data collection and valuable feedback.

\bibliography{anthology,custom}
\bibliographystyle{acl_natbib}

\appendix

\input{sections/6_appendix}

\end{document}

%% file: sections/0_abstract.tex
\begin{abstract}

In this paper, we introduce Cross-View Language Modeling, a simple and effective pre-training framework that unifies cross-lingual and cross-modal pre-training with shared architectures and objectives. Our approach is motivated by a key observation that cross-lingual and cross-modal pre-training share the same goal of aligning two different views of the same object into a common semantic space. To this end, the cross-view language modeling framework considers both multi-modal data (i.e., image-caption pairs) and multi-lingual data (i.e., parallel sentence pairs) as two different views of the same object, and trains the model to align the two views by maximizing the mutual information between them with conditional masked language modeling and contrastive learning. We pre-train \textbf{\baby}, a \textbf{C}ross-lingual \textbf{C}ross-modal \textbf{L}anguage \textbf{M}odel, with the cross-view language modeling framework. Empirical results on IGLUE, a multi-lingual multi-modal benchmark, and two multi-lingual image-text retrieval datasets show that while conceptually simpler, \baby significantly outperforms the prior state-of-the-art with an average absolute improvement of over 10\%. Moreover, \baby is the first multi-lingual multi-modal pre-trained model that surpasses the translate-test performance of representative English vision-language models by zero-shot cross-lingual transfer.\footnote{The code and pre-trained models are available at \url{https://github.com/zengyan-97/CCLM}.}

\end{abstract}

%% file: sections/1_introduction.tex
\section{Introduction}

Recently, the tremendous success of self-supervised language model pre-training~\citep{peters2018elmo,radford2018improving,devlin2018bert,liu2019roberta,radford2019language,DBLP:conf/nips/00040WWLWGZH19,raffel2019exploring,lewis2019bart,brown2020language} has been expanded to the multi-lingual~\citep{lample2019cross,conneau2019unsupervised,pfeiffer2020mad,chi2020infoxlm} and multi-modal~\citep{lu2019vilbert,tan2019lxmert,su2019vl,chen2020uniter,li2020oscar} domain. Advances on multi-lingual pre-training enables cutting-edge language technology to benefit a much boarder group of users including non-English speakers. Similarly, multi-modal pre-training makes pre-trained models applicable to a much larger set of tasks and user groups. Both of these directions make people's lives in a multi-lingual multi-modal world easier. Therefore, a natural next step is to explore multi-lingual multi-modal pre-training which enables pre-trained models to solve multi-modal tasks expressed in non-English languages without the need of collecting training data in these languages, which can be very costly for certain low-resource languages.

While appealing, multi-lingual multi-modal pre-training has its own challenges. Unlike multi-lingual pre-training and multi-modal pre-training where relatively large amount of parallel data is available, there exists only a few multi-lingual multi-modal corpora and their language coverage is also limited. Two pioneering works, M$^3$P~\citep{ni2021m3p} and UC$^2$~\citep{zhou2021uc2}, propose to pivot either on English texts or images to align multi-lingual multi-modal representations. Both of them introduce a number of new objectives to make use of the anchor for alignment. However, a recent benchmark on multi-lingual multi-modal pre-training ~\citep{bugliarello2022iglue} reveals that these multi-lingual multi-modal pre-trained models are still falling short: while achieving seemingly promising zero-shot cross-lingual transfer performance on some vision-and-language tasks, they still significantly under-perform ``translate-test'', a simple baseline which translates the test examples into English and uses an English-only vision-language model for inference. This prevents existing multi-lingual multi-modal models to be applicable in real-world applications. In contrast, multi-lingual pre-trained models such as XLM-R~\citep{conneau2019unsupervised} significantly outperforms the translate-test baseline in most languages and is widely used in practical applications.

This paper aims to fully exploit the potential of multi-lingual multi-modal pre-training. We point out two major limitation of current state-of-the-arts. First, existing methods do not exploit parallel text corpora, which can be easily collected and are abundant for many language pairs. Instead, M$^3$P performs masked language modeling with monolingual texts in different languages for multi-lingual alignment. However, parallel texts are shown to be more helpful according to multi-lingual pre-training literature~\citep{conneau2019unsupervised,chi2020infoxlm}. Second, a number of new pre-training objectives involving specific architecture changes and different input-output formats are introduced for English or image pivoting, making it non-trivial to combine them together for better performance and scale to larger data.

In this work, we argue that multi-lingual and multi-modal pre-training are essentially achieving the same goal of aligning two different views of a same object into a common semantic space. Therefore, we believe these two seemingly different strategies can be combined into a unified framework. To this end, we introduce cross-view language modeling, a simple and effective framework that unifies cross-lingual and cross-modal pre-training with shared architecture and objectives. Specifically, we consider both multi-modal data (i.e., image-caption pairs) and multi-lingual data (i.e., parallel sentence pairs) as pairs of two different views of the same object. With either multi-modal or multi-lingual data as input, we encode the two views with Transformer models and then fuse their representations with a cross-attention Transformer model shared for both cross-modal and cross-lingual fusion. We train the model to align the two views into a common semantic space by maximizing the mutual information between them with a conditional masked language modeling objective, a contrastive learning objective, and a matching objective. In this way, the cross-view language modeling framework unifies English pivoting and image pivoting schemes seamlessly and makes the best of both worlds.

To evaluate the effectiveness of our approach, we pre-train \baby, a Cross-lingual Cross-modal Language Model, with the proposed cross-view language modeling framework. Experimental results show that \baby significantly outperforms prior state-of-the-art with an averaged absolute improvement of over 10\% and 30\% on multi-lingual vision-language understanding and retrieval tasks in terms of accuracy and R@1 on IGLUE~\citep{bugliarello2022iglue}, a recently released multi-lingual multi-modal benchmark. Notably, \baby is the first multi-lingual vision-language model that surpasses the ``translate-test'' performance of mono-lingual vision-language models via zero-shot cross-lingual transfer, which we believe is a crucial step towards practical multi-lingual multi-modal pre-training. Since previous work used different pre-training datasets, making direct comparison difficult, we also conduct an in-depth ablation study to investigate the contribution of different parts in our framework. The results show that use of parallel sentence pairs helps to fully exploit the potential of language pivoting for multi-lingual multi-modal pre-training and also confirm the importance of unified architectures and objectives in \baby.

\textbf{Contributions. (1)} We propose a cross-view language modeling framework that unifies multi-lingual and multi-modal pre-training with shared architectures and objectives. 
\textbf{(2)} \baby advances the state-of-the-art of multi-lingual vision-language pre-training by a large margin. It also surpasses the translate-test baseline for the first time, demonstrating the potential of multi-lingual multi-modal pre-training. \textbf{(3)} We further scale up \baby with massive pre-training data and larger model size. We will release our large-scale pre-trained multi-lingual multi-modal models to benefit a larger set of tasks and user groups and setup a strong and easily reproducible baseline for multi-lingual multi-modal research.

%% file: sections/2_related.tex
\section{Related Work}

\paragraph{Multi-lingual Pre-training}

Multilingual BERT~\citep{devlin2018bert} demonstrates that good cross-lingual transfer results can be achieved by performing masked language modeling on multi-lingual corpora with shared vocabulary and weight. Later, XLM~\citep{lample2019cross}, XLM-R~\citep{conneau2019unsupervised}, and Unicoder~\citep{huang2019unicoder} introduce a number of new objectives including translation language modeling (TLM), cross-lingual word recovery, and cross-lingual paraphrase classification to improve multi-lingual pre-training. More recently, MAD-X~\citep{pfeiffer2020mad} and InfoXLM~\citep{chi2020infoxlm} further improve multi-lingual pre-training via adapter~\citep{houlsby2019parameter} and contrastive learning.

\paragraph{Vision-Language Pre-training}

Inspired by the success of language model pre-training, a number of work~\citep{lu2019vilbert,tan2019lxmert, li2020oscar,chen2020uniter,zeng2021multi, wang2022ofa, yu2022coca} investigates vision-language pre-training on large scale image-caption pairs and proposes a number of objectives to align vision and language representations, including masked multi-modal modeling, multi-modal alignment prediction, RoI feature regression, image-text matching, to name a few. Vision-language pre-training has reshaped the landscape of vision-and-language research and pushed the state-of-the-arts on a wide range of vision-language tasks~\citep{zhou2022vlue}. However, it is non-trivial to collect large scale image-caption pairs in other languages. As such, most existing vision-language pre-trained models are limited to English tasks.

\paragraph{Multi-lingual Multi-modal Pre-training}
Multi-lingual multi-modal pre-training aims to make multi-modal models applicable on non-English texts by cross-lingual transfer. In this paper we mainly consider multi-modal in the vision-language context. 
The key difficulty of multi-lingual multi-modal pre-training is the lack of non-English image-text pairs.
Two representative works tackle the lack of non-English image-text pairs by pivoting on either English texts or images. Specifically, M$^3$P~\citep{ni2021m3p} uses English as pivot and alternates between English-only vision-language pre-training and multi-lingual masked language modeling. UC$^2$~\citep{zhou2021uc2}, on the other hand, translates English captions into multiple languages and considers images as the anchor, achieving state-of-the-art on various multi-lingual vision-language tasks. More recently, MURAL~\citep{jain2021mural} collects large-scale image-text pairs in 110 languages and pre-trains a dual encoder model via contrastive learning. MURAL achieves new state-of-the-art on multi-lingual image-text retrieval tasks. However, the dual encoder architecture of MURAL makes it unable to perform multi-modal understanding tasks well.

%% file: sections/3_method.tex
\section{Cross-View Language Modeling}
\begin{figure*}[t]
\begin{center}
\centerline{\includegraphics[width=1.8\columnwidth]{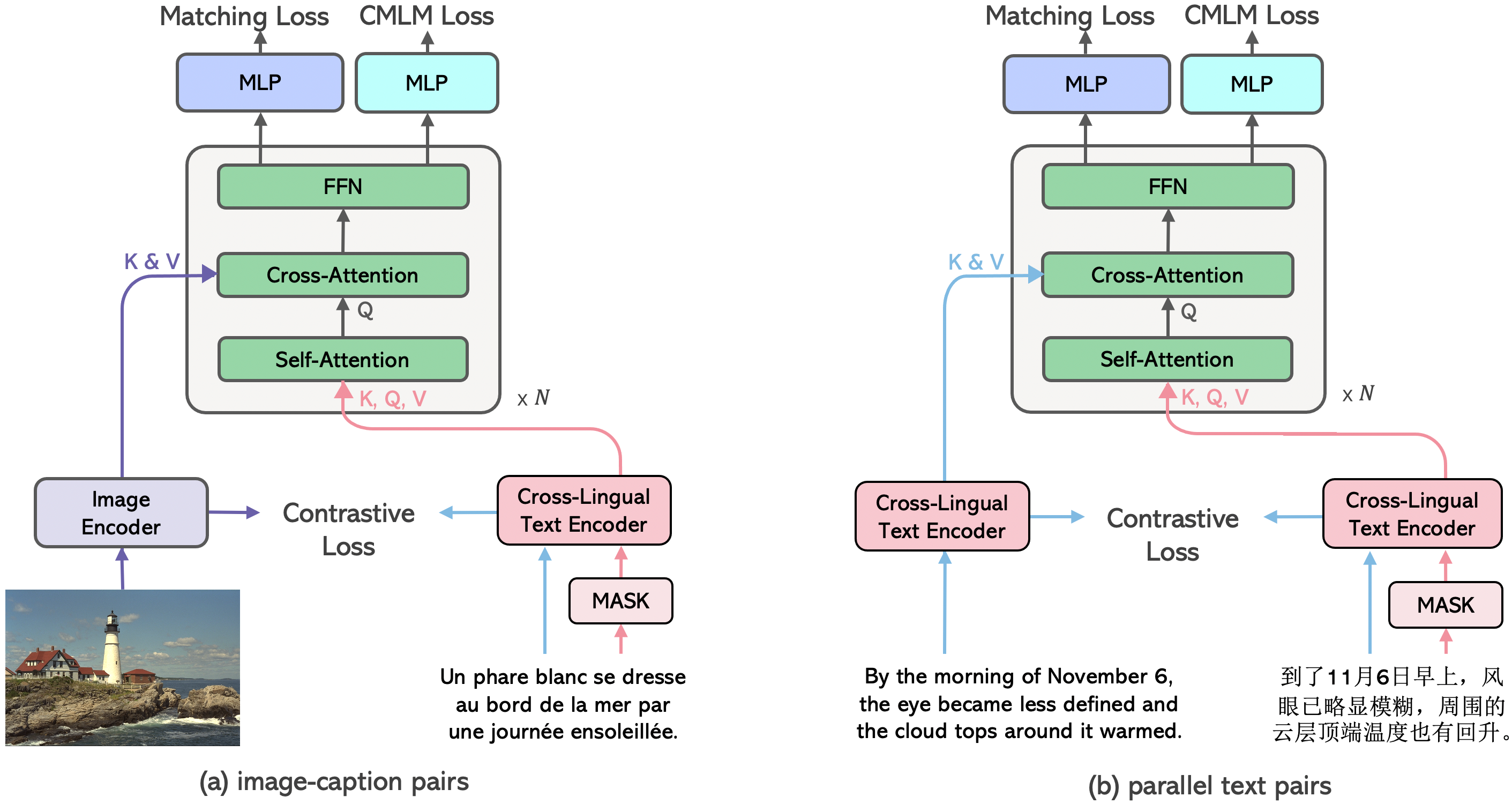}}
\caption{Illustration of the cross-view language modeling framework. \baby takes two different views of the same object, i.e., either (a) image-caption pairs or (b) parallel sentence pairs, as input. \baby first encodes the two views separately. Then the representations of the two views are fused by a Transformer-based model, which is shared for both cross-lingual and cross-modal fusion. \baby is optimized by maximizing the mutual information between the two views via conditional masked language modeling loss, contrastive loss, and matching loss. 
}
\label{Fig:model}
\end{center}
\vspace{-0.5cm}
\end{figure*}

\subsection{Overview}
Cross-view language modeling is a simple framework that unifies cross-lingual pre-training and cross-modal pre-training with shared architecture and objectives. \baby consists of an image encoder, a cross-lingual text encoder, and a fusion model. All components are Transformer-based. Specifically, the image encoder~\cite{dosovitskiy2020image} first splits an image into non-overlapping patches, and then embeds these patches with transformer layers, yielding $\{\Vec{v}_\mathrm{cls}, \Vec{v}_1,...,\Vec{v}_{N_1}\}$. For an image of resolution of 224x224 and patch size of 32x32, we have $N_1=49$. Similarly, the cross-lingual text encoder encodes a text input via transformer layers, yielding $\{\Vec{w}_\mathrm{cls}, \Vec{w}_1,...,\Vec{w}_{N_2}\}$. $N_2$ is the length of the text input. Then, the fusion model fuses text features with the corresponding image features or features of the translated text based on cross-attention, producing $\{\Vec{x}_\mathrm{cls}, \Vec{x}_1,...,\Vec{x}_{N_2}\}$.

As illustrated in Figure~\ref{Fig:model}, with either (text, image) pairs or (text, translation) pairs as input, we consider the paired input as two different views and train the model to align their representations in a common semantic space. This unified cross-view perspective allows us to share input-output formats, architectures, and training objectives between cross-lingual inputs and cross-modal inputs. Specifically, we completely share the fusion model for both cross-lingual fusion and cross-modal fusion, and optimize the model by contrastive loss, matching loss, and conditional masked language modeling loss for both cross-lingual and cross-modal inputs. We select these objectives because they are universally effective in both cross-lingual and cross-modal pre-training literature~\citep{chi2020infoxlm,li2021align}. We will show that the three loss maximize sequence-level and token-level mutual information between image-caption pairs or parallel sentence pairs. On the other hand, we empirically find that the three loss are more effective for cross-lingual cross-modal pre-training than certain task-specific loss such as masked region-to-token language modeling which is specially for multi-modal pre-training or translation language modeling for multilingual pre-training.

\subsection{A Mutual Information Maximization Perspective}

In this section, we explain our approach from an information-theoretic perspective. Formally, given two random variables $A$ and $B$, mutual information $I(A, B)$ measures dependencies between the two random variables. We define $A=a$ and $B=b$ as two different views of a data point, which can be either an image-caption pair or a parallel sentence pair. In this case, we will show that \baby maximizes a lower bound of $I(A, B)$ for cross-lingual cross-modal pre-training by minimizing the InfoNCE loss \cite{oord2018representation} defined as:

\begin{small} 
\begin{equation}
\mathcal{L}_\mathrm{nce} = -\mathbb{E}_{p(A,B)}\left[ \log \frac{\exp (f_{\bm{\theta}}(a,b))}{  \sum_{\tilde{b}\in\tilde{B}} \exp (f_{\bm{\theta}}(a,\tilde{b})) }  \right],
\end{equation}
\end{small}

where $f_{\bm{\theta}} \in \mathbb{R}$ is a function parameterized by
$\bm{\theta}$ and $\mathcal{\tilde{B}}$ contains the positive sample $b$ and $|\mathcal{\tilde{B}}| - 1$ negative samples.

The contrastive loss between the image encoder and the cross-lingual text encoder is a symmetric version of $\mathcal{L}_\mathrm{nce}$:

\begin{small} 
\begin{align}
\mathcal{L}_\mathrm{cl} = -\frac{1}{2} \mathbb{E}_{p(A,B)} & \big[ \log \frac{\exp (f_{\bm{\theta}}(a,b))}{\sum_{\tilde{b}\in\tilde{B}} \exp (f_{\bm{\theta}}(a,\tilde{b}))} \nonumber \\
& + \log \frac{\exp (f_{\bm{\theta}}(a,b))}{\sum_{\tilde{a}\in\tilde{A}} \exp (f_{\bm{\theta}}(\tilde{a},b))}\big], 
\end{align}
\end{small}

where $|\tilde{A}|=|\tilde{B}|=N$ is the batch size, and we predict $(a,b)$ pairs from in-batch negatives. $f_{\bm{\theta}}(a,b) = g_v(\Vec{v}_\mathrm{cls})^\top g_w(\Vec{w}_\mathrm{cls}) / \tau$ given an image-caption pair or $f_{\bm{\theta}}(a,b) = g_w(\Vec{w}^a_\mathrm{cls})^\top g_w(\Vec{w}^b_\mathrm{cls}) / \tau $ given a translation pair. $\Vec{v}_\mathrm{cls}$ and $\Vec{w}_\mathrm{cls}$ are the output \texttt{[CLS]} embedding of the image encoder \footnote{Some vision transformers, e.g. Swin-Transformer, use the output of average pooling layer as the \texttt{[CLS]} embedding.} and the cross-lingual text encoder. $g_v$ and $g_w$ are transformations that map the \texttt{[CLS]} embeddings to normalized lower-dimensional representations. $\tau$ is a learnable temperature parameter.

Similarly, the matching loss applied on the output \texttt{[CLS]} embedding of the fusion model (denoted as $\Vec{x}_\mathrm{cls}(a, b)$) can also be viewed as a symmetric version of $\mathcal{L}_\mathrm{nce}$:

\begin{scriptsize} 
\begin{align}
\mathcal{L}_\mathrm{match} = -\frac{1}{2} \mathbb{E}_{p(A,B)} & \big[ \log \frac{\exp (f_{\bm{\theta}}(a,b))}{\exp (f_{\bm{\theta}}(a,b)) +
\exp (f_{\bm{\theta}}(a,b_\mathrm{neg}))} \nonumber \\
& + \log \frac{\exp (f_{\bm{\theta}}(a,b))}{\exp (f_{\bm{\theta}}(a,b) + \exp  f_{\bm{\theta}}(a_\mathrm{neg},b))}\big],
\end{align}
\end{scriptsize}

where we only sample a negative instance for each ground-truth $(a, b)$ pair and predict whether a pair is matched (true or false). In this case, $f_{\bm{\theta}}(a,b)=\Vec{v}_\mathrm{true}^\top \Vec{x}_\mathrm{cls}(a, b)$, where $\Vec{v}_\mathrm{true}$ is a parametric vector. 

The conditional MLM loss can also be interpreted as maximizing mutual information~\cite{KongdYLDY20} between the context $c=(\hat{a}, b)$ ($\hat{a}$ denotes the masked text input, and $b$ is the corresponding image or translated text) and the masked token $w_i$ in $a$: 

\begin{small} 
\begin{equation}
\mathcal{L}_\mathrm{mlm} = - \mathbb{E}_{p(C,W)} \big[\log \frac{\exp (f_{\bm{\theta}}(c,w_i))}{  \sum_{\tilde{w}\in\mathcal{V}} \exp (f_{\bm{\theta}}(c,\tilde{w})) } \big], 
\end{equation}
\end{small}
where $f_{\bm{\theta}}(c, w_i) = \psi(w_i)^\top  \Vec{x}_i(\hat{a},b)$. $\Vec{x}_i$ is the output vector at $w_i$ position of the fusion model. $\psi(w): \mathcal{V} \rightarrow \mathbb{R}^d$ is a lookup function that maps a word token $w$ into a parametric vector. $\mathcal{V}$ is the full vocabulary set.

Finally, the pre-training objective of \baby is defined as: $\mathcal{L} = \mathcal{L}_\mathrm{cl} + \mathcal{L}_\mathrm{match} + \mathcal{L}_\mathrm{mlm}$,  where the contrastive loss and matching loss maximize sequence-level mutual information while the MLM loss maximizes token-level mutual information, which are complement of each other. 

%% file: sections/4_experiments.tex
\section{Experiment}

\subsection{Experimental Settings}

\subsubsection{Pre-training Datasets}
We pre-train \baby on the combination of image-caption pairs and parallel multilingual texts. Appendix~\ref{app:baselines} describes compared models in details.

\noindent\textbf{Multi-modal Data} For image-caption pairs, we follow the practice of UC$^2$ to make a fair comparison and use their released translation-augmented version of CC3M dataset. It contains the original CC3M image-caption pairs~\cite{sharma2018conceptual} and machine-translated captions in five different languages (German, French, Czech, Japanese, and Chinese). 
This multi-modal dataset is widely utilized by previous work, including UC$^2$, mUNITER and xUNITER. We denote this variant as CCLM$^\text{3M}$. In additional to this setting, we leverage large-scale vision language pre-training by utilizing the pre-trained weights of X$^2$-VLM~\cite{zeng-2021-multi, zeng2022x} which has been trained on more than 1B image-text pairs in English. Based on it we apply the proposed framework for multi-lingual multi-modal pre-training.

\noindent\textbf{Multi-lingual Data} Previous work such as mUNITER, xUNITER, and M$^3$P use large-scale monolingual texts in different languages, namely multi-lingual Wikipedia 101G dataset, for multilingual alignment. Differently, we propose to utilize parallel text corpus. We collect a subset of the WikiMatrix~\citep{schwenk2019wikimatrix} dataset containing parallel texts between English and other languages in the IGLUE benchmark. 
Appendix~\ref{app:data} shows the number of pairs per language. In total, the dataset consists of 19M parallel sentence pairs. 

\subsubsection{Implementation Details}

CCLM$_\mathrm{base}$ consists of 12 Transformer layers for the image encoder and the text encoder respectively. CCLM$_\mathrm{large}$ consists of 24 layers for each encoder. The fusion encoder contains 6 Transformer layers for both CCLM$_\mathrm{base}$ ($d=768$) and CCLM$_\mathrm{large}$ ($d=1024$). In total, CCLM$_\mathrm{base}$ and CCLM$_\mathrm{large}$ consist of $\sim420$M and $\sim970$M parameters respectively. Following existing models such as M$^3$P and UC$^2$, we also utilize XLM-R~\cite{conneau2019unsupervised} as the text encoder. Concretely, CCLM$^\text{3M}$ is initialized with a pre-trained image encoder~\cite{LiuL00W0LG21} and XLM-R. \baby is initialized with the pre-trained X$^2$-VLM~\cite{zeng-2021-multi, zeng2022x} and XLM-R. 

In pre-training, the image encoder takes images of resolution of $224\times224$ as input for pre-training. During fine-tuning, we increase the image resolution to $384\times384$ and interpolate the positional embeddings of image patches following \citet{dosovitskiy2020image}. The maximum sequence length is set to 30 and 64 for image captions and parallel multilingual texts respectively.  We apply mixed precision for pre-training. We use the AdamW~\cite{loshchilov2018decoupled} optimizer with a weight decay of 0.02. We mix different types of data in a training batch. Following UC$^2$, to make a fair comparison, we train CCLM$^\text{3M}$ for 30 epochs on 8 NVIDIA A100 GPUs and the batch size is set to 1024, which tasks $\sim1.5$ days. The learning rate is warmed-up to $1e^{-4}$ in the first 2500 steps and decayed linearly. We train \babyB and \babyL for 40 epochs.

\begin{table*}[t]
\begin{center}
\resizebox{0.75\linewidth}{!}{
\begin{tabular}{lrrcrrrr}
\toprule
\multirow{2}{*}{\small \shortstack{\textbf{Model}}}         &  \multicolumn{1}{c}{\textbf{NLI}}   & \multicolumn{1}{c}{\textbf{QA}} & \multicolumn{1}{c}{\textbf{Reasoning}}  & \multicolumn{4}{c}{\textbf{Retrieval}} \\
  \cmidrule(r){2-2}
  \cmidrule(r){3-3}
  \cmidrule(r){4-4}
  \cmidrule(r){5-8}
 & \multicolumn{1}{c}{XVNLI} & \multicolumn{1}{c}{xGQA}     & \multicolumn{1}{c}{MaRVL} & \multicolumn{2}{c}{xFlickr\&CO}          & \multicolumn{2}{c}{WIT}        \\
 & & & & \multicolumn{1}{c}{IR} & \multicolumn{1}{c}{TR} & \multicolumn{1}{c}{IR} & \multicolumn{1}{c}{TR} \\

\midrule
 \multicolumn{8}{c}{\emph{Translate everything to English and use English-only model (Translate-Test)
}} \\
\midrule

UNITER & 73.65 & \bf 50.62 & 61.92 & 41.04 & \bf 37.49 & \bf 15.43 & 16.01 \\
ViLBERT & 73.45 & 50.33 & 62.39 & 36.97 & 33.21 & 15.40 & \bf 16.93 \\
VisualBERT & \bf 74.12 & 48.72 & 62.35 & \bf 41.64 & 36.44 & 15.36 & 15.75 \\
VL-BERT & 73.86 & 49.78 & \bf 64.16 & 38.18 & 31.84 & 15.11 & 16.09 \\

\midrule
 \multicolumn{8}{c}{\emph{Fine-tune model on English training set (Zero-Shot)}} \\
\midrule
mUNITER & 53.69 & 9.97 & 53.72 & 8.06 & 8.86 & 9.16 & 10.48 \\
xUNITER & 58.48 & 21.72 & 54.59 & 14.04 & 13.51 & 8.72 & 9.81 \\
M$^3$P & 58.25 & 28.17 & 56.00 & 12.91 & 11.90 & 8.12 & 9.98 \\
UC$^2$ & 62.05 & 29.35 & 57.28 & 20.31 & 17.89 & 7.83 & 9.09 \\

\midrule

\babymmm & 74.64 & 42.36 & 65.91 & 67.35 & 65.37 & 27.46 & 28.66 \\  

\babyB & 74.78 & 48.12 & 68.49 & 76.94 & 76.22 & 33.90 & 35.26 \\ 
\babyL & \bf 78.95 & \bf 56.25 & \bf 74.83 & \bf 83.78 & \bf 83.46 & \bf 43.74 & \bf 44.88 \\ 

\bottomrule
\end{tabular}}
\end{center}
\caption[caption]{\textbf{Results on IGLUE benchmark.} 
R@1 and Accuracy are reported for retrieval tasks (xFlickr\&CO and WIT) and understanding tasks (XVNLI, xGQA, MaRVL) respectively. 
In the zero-shot setting, the models are fine-tuned on English train sets and directly evaluated on target languages. We report few-shot results in Appendix~\ref{app:iglue}. 
} 
\label{tab:iglue}
\end{table*}

\subsubsection{Downstream Tasks}

We evaluate \baby on the IGLUE benchmark~\cite{bugliarello2022iglue}, a recently released benchmark for evaluating multi-lingual multi-modal pre-training, and a multi-lingual image-text retrieval benchmark including the multi-lingual version of Flickr30K~\citep{young2014image,elliott2016multi30k} and MSCOCO~\citep{chen2015microsoft}. Note that \baby can also be applied on generation tasks such as image captioning by following the adaptation strategy of X-VLM~\cite{zeng2022x,zeng-nie-2021-investigation}.

\noindent\textbf{XVNLI}: The Cross-lingual Visual NLI dataset is collected by combining SNLI~\citep{bowman2015large} with its multi-modal~\citep{xie2019visual} and multi-lingual~\citep{agic2017baselines} counterparts. It requires the model to predict if a text-hypothesis ``entails'', ``contradicts'', or is
``neutral'' to an image-premise.

\noindent\textbf{xGQA}: The Cross-lingual Grounded Question Answering task~\citep{pfeiffer2021xgqa} is collected by manually translating the GQA~\citep{hudson2019gqa} validation set into 7 languages. It requires a model to answer several types of structured questions about an image. We model GQA as a generation task following \citet{li2021align}.

\noindent\textbf{MaRVL}: The Multicultural Reasoning over Vision and Language dataset~\citep{liu2021visually} requires to determine whether a textual description is true or false about a pair of images. The MaRVL dataset is used for testing and the NLVR2~\citep{suhr2018corpus} dataset is used for training.

\noindent\textbf{xFlickr\&CO and WIT}: The xFlickr\&CO dataset is collected by combining 1000 images from Flickr30K and MSCOCO respectively and crowdsource image descriptions in 6 other languages. Similarly, the Wikipedia-based Image Text dataset~\citep{srinivasan2021wit} is collected from Wikipedia in 108 languages. We follow the data preprocessing and splitting details in IGLUE for both datasets.

\noindent\textbf{Multi30K}: This dataset~\cite{elliott2016multi30k} extended Flickr30K~\citep{young2014image} from English (en) to German (de), French (fr) and Czech (cs). It contains 31,783 images and provides five captions per image in English and German, and one caption per image in French and Czech. Dataset splits are defined as the original Flickr30K.

\noindent\textbf{MSCOCO}: This dataset extends the MSCOCO caption dataset~\citep{chen2015microsoft} by translating the captions into Japanese~\citep{yoshikawa2017stair} and Chinese~\citep{li2019coco}. The Japanese and Chinese subsets consist of 820k and 20k captions respectively. Following previous work, we use the same train, dev, and test splits for English and Japanese as defined in \citet{karpathy2015deep}. As for Chinese, we use the COCO-CN split~\citep{li2019coco}. 

For all retrieval tasks, we follow previous work~\cite{li2021align} and X-VLM~\cite{zeng2021multi}. During fine-tuning, we optimize $\mathcal{L}_\mathrm{cl}$ and $\mathcal{L}_\mathrm{match}$. For inference, we first compute similarity for all images and texts, and then take the top-k candidates and calculate the final ranking scores using the fusion model.

\subsection{Experimental Results} 
\subsubsection{Results on IGLUE Benchmark}

Table~\ref{tab:iglue} shows \baby performance on the IGLUE benchmark. First, for zero-shot cross-lingual transfer, we can see that \babymmm outperforms all compared models by a substantial margin while pre-trained on the same multi-modal data. Specifically, compared to UC$^2$, the prior state-of-the-art, \babymmm obtains an average accuracy improvement of 11.4\% on multi-lingual multi-modal understanding tasks including XVNLI, xGQA, and MaRVL, and an average R@1 improvement of 47.3\% and 18.2\% on multi-lingual multi-modal retrieval datasets including xFlickr\&CO and WIT. This confirms that previous multi-lingual multi-modal models fail to fully exploit the potential of multi-lingual multi-modal pre-training and our proposed cross-view language modeling framework can better align multi-lingual multi-modal representations with unified objectives.

We also find that the performance of our framework can be significantly improved by leveraging large-scale image-text pre-training in English (\babyB) and/or scaling up the model size (\babyL). Notably, \baby is the first multi-lingual multi-modal pre-trained model that substantially outperforms the translate-test results of representative English VLMs tested in the IGLUE benchmark. This, for the first time, proves the potential of multi-lingual multi-modal pre-training on building practical real-world applications involving vision-language tasks in different languages.

\subsubsection{Results on Multi-lingual Retrieval}

\begin{table}[t]
\begin{center}
\resizebox{\linewidth}{!}{
\begin{tabular}{lccccccc}
\toprule
\multirow{2}{*}{\small \shortstack{\textbf{Model}}} & \multicolumn{4}{c}{\small \textbf{Multi30K}} & \multicolumn{3}{c}{\small \textbf{MSCOCO}} \\
\cmidrule(r){2-5}
\cmidrule(r){6-8}
  & EN  & DE & FR & CS & EN & ZH & JA  \\ 
\midrule
$\text{M}^3\text{P}$ & 87.7 & 82.7 & 73.9 & 72.2 & 88.7 & 86.2 & 87.9  \\
UC$^2$ & 88.2 & 84.5 & 83.9 & 81.2 & 88.1 & 89.8 & 87.5  \\ 
MURAL$_\mathrm{base}$ & 92.2 & 88.6 & 87.6 & 84.2 & 88.6 & - & 88.4 \\
MURAL$_\mathrm{large}$ & 93.8 & 90.4 & 89.9 & 87.1 & 92.3 & - & 91.6 \\
\midrule
\babymmm & 95.3 & 92.4 & 92.1 & 91.2 & 93.1 & 92.2 & 93.2 \\  
\babyB & 97.2 & 94.6 & 95.5 & 94.8 & 95.4 & 93.2 & 95.7 \\
\babyL & \bf 97.8 & \bf 95.8 & \bf 96.6 & \bf 96.2 & \bf 95.6 & \bf 94.0 & \bf 96.1 \\

\bottomrule
\end{tabular}}
\end{center}
\caption[caption]{\textbf{Results on multi-lingual image-text retrieval} in all-language fine-tune setting, where a model is fine-tuned on the combination of training data in all languages. Following previous work, we compute the average Recall@K for both image-to-text retrieval and text-to-image retrieval with K = 1, 5, 10, as the evaluation metric. 
We additionally report results in other fine-tune settings in Appendix~\ref{app:retrieval}.
}
\label{tab:retrieval}
\end{table}

Table~\ref{tab:retrieval} gives the results on the multi-lingual image-text retrieval benchmark. When pre-trained on the same multi-modal data, \babymmm substantially outperforms UC$^2$, the prior state-of-the-art, with an averaged improvement of over 10\% (in terms of averaged recall) across four languages on Multi30K. This confirms that our approach can better align multi-lingual multi-modal representations. \babymmm even outperforms MURAL. This is notable because MURAL$_\mathrm{large}$ is larger than our model and is pre-trained on much more data ($\sim 450 \times$ more image-text pairs and 390$\times$ more parallel sentence pairs). Moreover, we show that \baby also outperforms MURAL without fine-tuning in Appendix~\ref{app:retrieval}.

We also find that the cross-view language modeling framework yields better performance if leveraging large-scale pre-training on image-text pairs in English (\babyB) and/or scaling up the model size (\babyL), which is consistent with the experimental results on the IGLUE benchmark. It confirms that the proposed framework is scalable to both massive data and larger model size.

\subsubsection{Cross-lingual Transfer Gap}

\begin{figure}[ht]
\begin{center}
\includegraphics[width=0.45\textwidth]{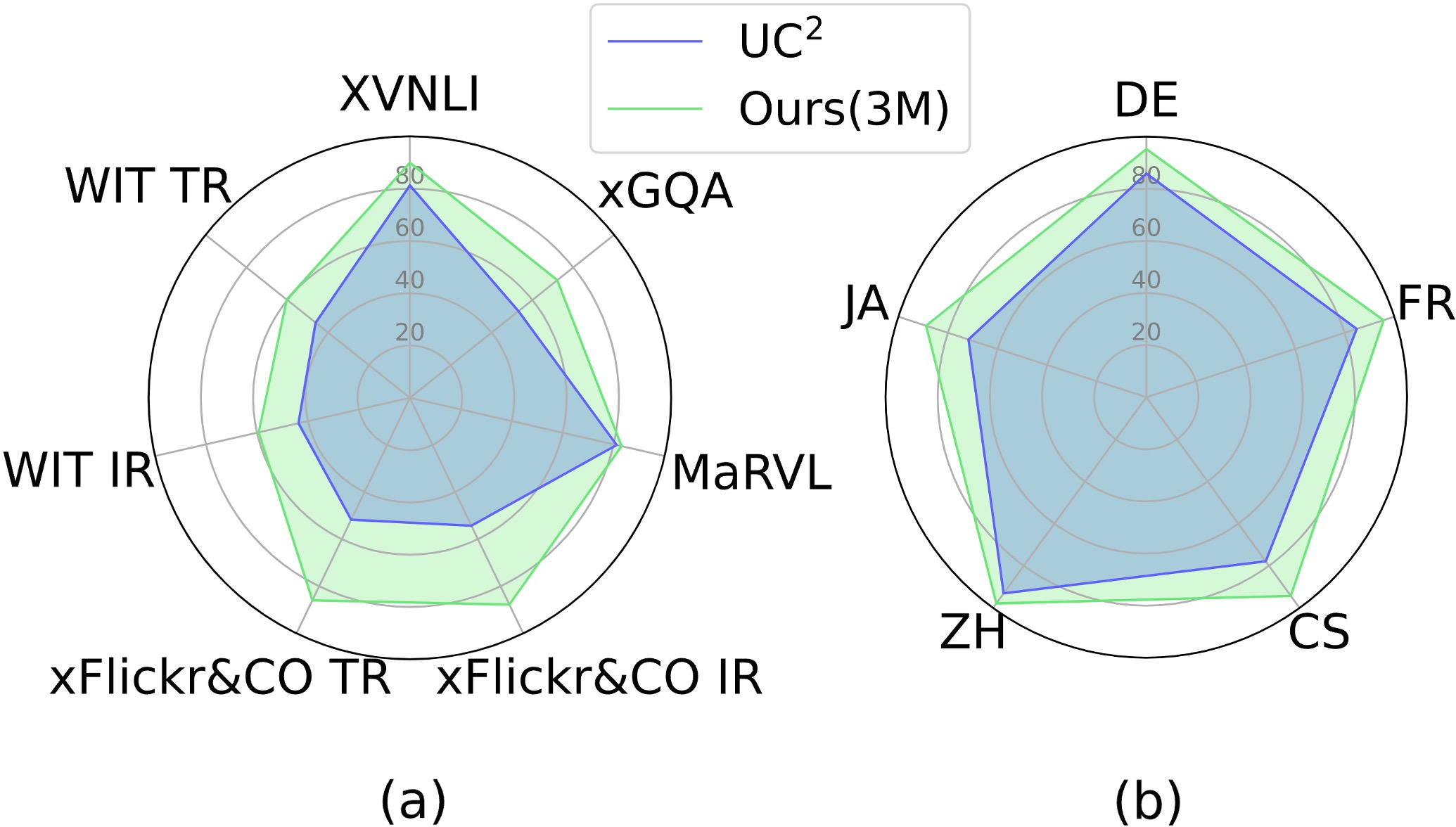}
\end{center}
\caption{Visualization of cross-lingual transfer gap.}
\label{Fig:transfer}
\end{figure}

\begin{table*}[t]
\begin{center}
\resizebox{0.75\linewidth}{!}{
\begin{tabular}{l|ccccc}
\toprule
\textbf{\multirow{2}*{Methods}} & \textbf{\multirow{2}*{Multi30K}} & \textbf{\multirow{2}*{MaRVL}} & \textbf{\multirow{2}*{xGQA}} & \multicolumn{2}{c}{\textbf{xFlickr\&CO}} \\
~ & ~ & ~ & ~ & IR & TR  \\
\midrule
Ours & \textbf{92.67} & \textbf{67.05} & \textbf{41.66} & \textbf{63.77} & \textbf{62.13} \\ 
\ -w/o shared cross-attn  & 92.49 & 66.67 & 36.76 & 63.73 & 62.01 \\  
\ -w/o shared FFN & 92.24 & 63.63 & 35.53 & 63.15 & 61.04 \\ 
\ -w/ TLM & 91.88 & 62.65 & 35.84 & 58.44 & 56.73 \\ 
\ -w/ TLM + CL & 92.34 & 65.00 & 36.13 & 63.42 & 61.33 \\ 
\ -w/o parallel sentence pairs & 91.90 & 58.37 & 28.80 & 44.11 & 43.24 \\ 
\bottomrule
\end{tabular}}
\end{center}
\caption[caption]{\textbf{Ablation study results.} Models w/o shared cross-attention and FFN are ablated variants where these modules are separately parameterized in the cross-lingual fusion model and the cross-modal fusion model. Models w/ TLM and TLM + CL are variants where the multi-lingual objectives are that used in XLM-R and InfoXLM, which are not unified with the multi-modal objectives. All compared models are pre-trained for 15 epochs.} 
\label{tab:ablation}
\end{table*}

In addition to absolute cross-lingual transfer results reported in Table~\ref{tab:iglue} and Table~\ref{tab:retrieval}, we also compare the cross-lingual transfer gap of different models. We visualize the ratio of a model's performance on non-English languages to its performance on English test set, in Figure \ref{Fig:transfer}. A larger radar chat indicates the model has a smaller relative transfer gap and can better transfer its performance to non-English test sets. We can see that \baby's relative cross-lingual transfer gap is consistently smaller than that of UC$^2$ across all tasks in the IGLUE benchmark (a) and all languages in the multi-lingual retrieval datasets (b). The absolute cross-lingual transfer gap is even more significant. For example, in Appendix~\ref{app:retrieval}, we can see that for M$^3$P, the absolute zero-shot cross-lingual transfer gap between EN-CS and EN-JA in Multi30K and MSCOCO are 41.4\% and 32.6\% respectively. This indicates that masked language modeling on unpaired texts in multiple languages are not very effective for cross-lingual alignment of multi-modal models. The gap for UC$^2$ is reduced to 13.2\% and 16.4\%, demonstrating the effectiveness of using machine-translated captions for multi-lingual multi-modal pre-training. \babymmm further reduces this gap to 5.4\% and 4.4\%. This confirms that the proposed cross-view language modeling framework can effectively transfer multi-modal representations from English to other languages without language-specific fine-tuning. In addition, we also visualize the multi-lingual text representations and image representations in \baby and a baseline approach in Appendix~\ref{app:vis}, which clearly shows our approach can better align multi-lingual image-text representations. 


\subsection{Ablation Study} 
Since previous work such as M$^3$P, UC$^2$, and MURAL all use different pre-training datasets, making direct comparison difficult, we conduct an in-depth ablation study to investigate the contribution of different design choices in the cross-view language modeling framework. We pre-train 5 ablated variants of \baby where parallel sentence pairs, unified architecture, or unified objectives are ablated. All compared models are pre-trained with the same CC3M and WikiMatrix data (except that w/o parallel sentence pairs) for 15 epochs to ensure a fair comparison. The results are shown in Table \ref{tab:ablation}.

First, we find that the use of parallel sentence pairs plays a very important role. This indicates that previous methods fail to fully exploit the potential of language pivoting for multi-lingual multi-modal pre-training. On the other hand, \baby variant trained without parallel sentences in Table~\ref{tab:ablation} which uses the same pre-training dataset as UC$^2$ still significantly outperforms previous models such as M$^3$P and UC$^2$.

We then compare other ablated variants which all utilized parallel sentence pairs. We find that separate parameterization of cross-attention and FFN modules for the cross-lingual and the cross-modal task in the fusion model leads to inferior results, especially for multi-lingual multi-modal understanding tasks such as xGQA.

Moreover, we conduct ablation study on loss functions. We mainly consider multi-lingual objectives because the multi-modal objective combination of itc+mlm+itm is the de-facto choice for multi-modal loss~\cite{li2021align, zeng2021multi}. We find that using common objectives in the multi-lingual pre-training literature underperforms our unified objective. These observations confirm the importance of unifying architectures and objectives for multi-lingual multi-modal pre-training.

%% file: sections/5_conclusion.tex
\section{Conclusion}

In this paper, we introduce cross-view language modeling, a simple and effective framework that unifies cross-lingual and cross-modal pre-training. It considers cross-lingual and cross-modal pre-training as the same procedure of aligning the representation of two different views of the same object, thus using shared model architectures and training objectives for multi-lingual multi-modal pre-training. We train \baby with the proposed framework and show that it advances the state-of-the-art on all downstream multi-lingual vision-language tasks by a large margin. Moreover, it surpasses the translate-test baseline for the first time, demonstrating the potential of multi-lingual multi-modal pre-training. Furthermore, the experimental results also confirm that the proposed framework is scalable to massive data and larger model sizes. We believe our model will become a foundation for future multi-lingual multi-modal research and serve as a strong baseline. Moreover, the cross-view language modeling framework also has the potential of unifying more modalities such as audio and video with the same architectures and objectives. We leave this for future work.

%% file: sections/6_appendix.tex
\clearpage

\section{Appendix}

\subsection{Compared Models}
\label{app:baselines}

\textbf{mUNITER and xUNITER:} A multi-lingual variant of the UNITER~\citep{chen2020uniter} model pre-trained by \citet{liu2021visually}. The model is pre-trained by alternating between a batch of multi-modal English data from CC3M with UNITER objectives and a batch of text-only multilingual Wikipedia data with the MLM objective. mUNITER and xUNITER differ in their initialization: mUNITER and xUNITER are initialized from mBERT and XLM-R.

\textbf{M$^3$P:} A multi-lingual multi-modal model initialized from XLM-R and pre-trained with the combination of multilingual masked language modeling, multi-modal code-switched masked language modeling, multi-modal code-switched masked region modeling, and multi-modal code-switched visual-linguistic matching. The code-switched training method allows the model to explicitly align images with non-English languages. In each multi-modal batch, image-text pairs are fed to the model either fully in English or with code-switched words according to a given sampling ratio. Similar to mUNITER and xUNITER, the model is trained by alternating multi-modal and multi-lingual batches.

\textbf{UC$^2$:} The state-of-the-art multi-lingual vision-language model which relies on (text-only) machine translation technologies to obtain CC3M data in five languages (Czech, French, German, Japanese, and Mandarin). The model is then pre-trained on multi-lingual multi-modal batches where a caption is sampled uniformly from the available languages for each image. As for pre-training objectives. In addition to conventional vision-language pre-training objectives, a visual-conditioned translation language modeling objective is added to improve multi-lingual multi-modal alignment.

\subsection{Details for Multi-lingual Data}
\label{app:data}

\begin{table}[h]
\begin{center}
\resizebox{\linewidth}{!}{
\begin{tabular}{cccccccccc}
\toprule
ES & FR & PT & RU & DE & VI & ID & AR & JA & ZH \\
\midrule
3,130 & 2,645 & 2,322 & 1,598 & 1,467 & 998 & 974 & 968 & 841 & 783 \\
\toprule
EL & CS & TR & DA & BG & KO & BN & ET  & TA & SW  \\
\midrule
609 & 509 & 455 & 412 & 353 & 281 & 269 & 241 & 61 & 51  \\
\bottomrule
\end{tabular}
}
\end{center}
\caption[caption]{The number of parallel sentence pairs per language (K) in the subset of WikiMatrix.}
\label{tab:wikimatrix}
\end{table}

\subsection{Results on English Tasks}
Table \ref{tab:entasks} reports \baby performance that is pre-trained on COCO, VG, SBU, and CC3M, on three common English multi-modal tasks. We can observe that \baby also has very competitive performance compared to strong English multi-modal baselines. 

\begin{table}[h]
\begin{center}
\resizebox{\linewidth}{!}{
\begin{tabular}{l|ccccc}
\toprule
\textbf{\multirow{2}*{Methods}} & \textbf{VQA2.0} & \multicolumn{2}{c}{\textbf{NLVR2}} & \multicolumn{2}{c}{\textbf{MSCOCO}(5K)} \\
~ & test-dev & dev  & test-P & IR & TR  \\
\midrule
VinVL$_\mathrm{base}$ & 75.95 & 82.05 & 83.08 & 58.10 & 74.60 \\
ALBEF (4M) & 74.54 & 80.24 & 80.50 & 56.80 & 73.10 \\
CCLM$_\mathrm{base}^\text{4M}$ & 77.17 & 82.66 & 83.22 & 60.89 & 77.72 \\
\bottomrule
\end{tabular}
}
\end{center}
\caption[caption]{\textbf{Results on common English multi-modal tasks.} 
R@1 and Accuracy are reported for MSCOCO (5K test set) and understanding tasks respectively. }
\label{tab:entasks}
\end{table}

\subsection{Few-Shot Results on IGLUE}
\label{app:iglue}

\begin{table*}[t]
\begin{footnotesize}
\begin{center}
\resizebox{0.8\linewidth}{!}{
\begin{tabular}{lrrcrrrr}
\toprule
\multirow{2}{*}{\small \shortstack{\textbf{Model}}}         &  \multicolumn{1}{c}{\textbf{NLI}}   & \multicolumn{1}{c}{\textbf{QA}} & \multicolumn{1}{c}{\textbf{Reasoning}}  & \multicolumn{4}{c}{\textbf{Retrieval}} \\
  \cmidrule(r){2-2}
  \cmidrule(r){3-3}
  \cmidrule(r){4-4}
  \cmidrule(r){5-8}
 & \multicolumn{1}{c}{XVNLI} & \multicolumn{1}{c}{xGQA}     & \multicolumn{1}{c}{MaRVL} & \multicolumn{2}{c}{xFlickr\&CO}          & \multicolumn{2}{c}{WIT}        \\
 & & & & \multicolumn{1}{c}{IR} & \multicolumn{1}{c}{TR} & \multicolumn{1}{c}{IR} & \multicolumn{1}{c}{TR} \\



\midrule
 \multicolumn{8}{c}{\emph{Few-shot train English fine-tuned model on target languages (Few-Shot) }} \\
\midrule
mUNITER & 53.95 & 37.21 & 53.41 & 8.54 & 9.32 & - & - \\
xUNITER & 60.55 & 40.68 & 57.46 & 14.30 & 13.54 & - & - \\
M$^3$P & 59.36 & 41.04 & 49.79 & 13.21 & 12.26 & - & - \\
UC$^2$ & 63.68 & 42.95 & 58.32 & 19.79 & 17.59 & - & - \\

CCLM$_\mathrm{base}^\text{3M}$ & \bf 75.15(.03) & \bf 50.94(.02) & \bf 70.53(.18) & \bf 66.04(.05) & \bf 68.15(.04) & - & - \\




\bottomrule
\end{tabular}}
\end{center}
\caption[caption]{\textbf{Few-Shot Results on IGLUE benchmark.} 
R@1 and Accuracy are reported for retrieval tasks (xFlickr\&CO and WIT) and understanding tasks (XVNLI, xGQA, MaRVL) respectively. For our model, mean and standard deviation (in brackets) of 3 different runs with different random seeds are reported.} 
\label{app:tab:iglue}
\end{footnotesize}
\end{table*}

Table~\ref{app:tab:iglue} gives results on IGLUE benchmark. For our models, mean and standard deviation (in brackets) of 3 different runs with different random seeds are reported. Results of compared models are directly copied from the IGLUE benchmark. In the few-shot setting, the English trained models are continually fine-tuned with a few labeled examples in a target language before evaluating on this language. We select exactly the same few-shot examples following IGLUE instructions to ensure our results are compatible with that reported in IGLUE. We omit few-shot evaluation on the WIT dataset because this setup is also omitted in IGLUE. We find that similar to existing models, \baby can also benefit from few-shot learning with a few examples in the target languages.

\subsection{More Results on Retrieval Tasks}
\label{app:retrieval}

\begin{table*}[t]
\begin{center}
\resizebox{0.8\linewidth}{!}{
\begin{tabular}{lccccccc}
\toprule
\multirow{2}{*}{\small \shortstack{\textbf{Model}}} & \multicolumn{4}{c}{\small \textbf{Multi30K}} & \multicolumn{3}{c}{\small \textbf{MSCOCO}} \\
\cmidrule(r){2-5}
\cmidrule(r){6-8}
  & EN  & DE & FR & CS & EN & ZH & JA  \\ 
\midrule
\multicolumn{8}{c}{\textit{English-only Fine-tune (Zero-Shot)}}\\
\midrule
$\text{M}^3\text{P}$ & 87.4 & 58.5 & 46.0 & 36.8 & 88.6 & 53.8 & 56.0  \\ 
UC$^2$ & 87.2 & 74.9 & 74.0 & 67.9 & 88.1 & 82.0 & 71.7  \\
\babymmm & \bf 94.8(.11) & \bf 90.3(.08) & \bf 90.9(.38) & \bf 89.4(.21) & \bf 93.2(.05) & \bf 91.0(.18) & \bf 88.8(.06) \\  
\midrule
\multicolumn{8}{c}{\textit{Single-Language Fine-tune}}\\
\midrule
$\text{M}^3\text{P}$ & 87.4 & 82.1 & 67.3 & 65.0 & 88.6 &75.8 &  80.1 \\ 
UC$^2$ & 87.2 &  83.8 &  77.6 &  74.2 & 88.1 &  84.9 & 87.3  \\
\babymmm & \bf 94.8(.11) & \bf 91.9(.16) & \bf 90.6(.18) & \bf 88.9(.05) & \bf 93.2(.05) & \bf 90.2(.24) & \bf 93.3(.26) \\  

\bottomrule
\end{tabular}}
\end{center}
\caption[caption]{\textbf{Results on multi-lingual image-text retrieval.} We compute the average Recall@K for both image-to-text retrieval and text-to-image retrieval with K = 1, 5, 10, as the evaluation metric. For our model, mean and standard deviation (in brackets) of 3 different runs with different random seeds are reported. 
}
\label{app:tab:retrieval}
\end{table*}

\begin{table}[h]
\begin{center}
\resizebox{\linewidth}{!}{
\begin{tabular}{lccccccc}
\toprule
\multirow{2}{*}{\small \shortstack{\textbf{Model}}} & \multicolumn{4}{c}{\small \textbf{Multi30K}} & \multicolumn{3}{c}{\small \textbf{MSCOCO}} \\
\cmidrule(r){2-5}
\cmidrule(r){6-8}
  & EN  & DE & FR & CS & EN & ZH & JA  \\ 
\midrule
MURAL$_\mathrm{base}$ &  82.4 & 76.2 & 75.0 & 64.6 & 79.2 & - & 73.4 \\
\babymmm & \bf 83.7 & \bf 79.1 & \bf 76.7 & \bf 73.9 & \bf 81.5 & \bf 79.5 & \bf 76.8 \\
\bottomrule
\end{tabular}
}
\end{center}
\caption[caption]{\textbf{Zero-shot results on multi-lingual image-text retrieval.} We compute the average Recall@K for both image-to-text retrieval and text-to-image retrieval with K = 1, 5, 10, as the evaluation metric. Results of compared models are directly copied from the corresponding papers.}
\label{tab:zeroretrieval}
\end{table}

Table~\ref{app:tab:retrieval} reports results on multi-lingual image-text retrieval of \baby. We follow the practice of prior work and evaluate in three different settings including English-only fine-tuning, single-language fine-tuning, and all-language fine-tuning, where the model is fine-tuned on English data, target language data, and the combination of training data in all languages, respectively.

We also report multi-lingual image-text retrieval results without fine-tuning (zero-shot) in Table~\ref{tab:zeroretrieval}. M$^3$P and UC$^2$ do not report their zero-shot retrieval performances. We can observe that \babymmm outperforms MURAL which is pre-trained on much larger data. Besides, the performance gap on non-English test sets of MURAL is larger, which shows our model has better cross-lingual transfer ability.



\subsection{Visualization of Representations}
\label{app:vis}

\begin{figure}[h]
\begin{center}
\centerline{\includegraphics[width=\columnwidth]{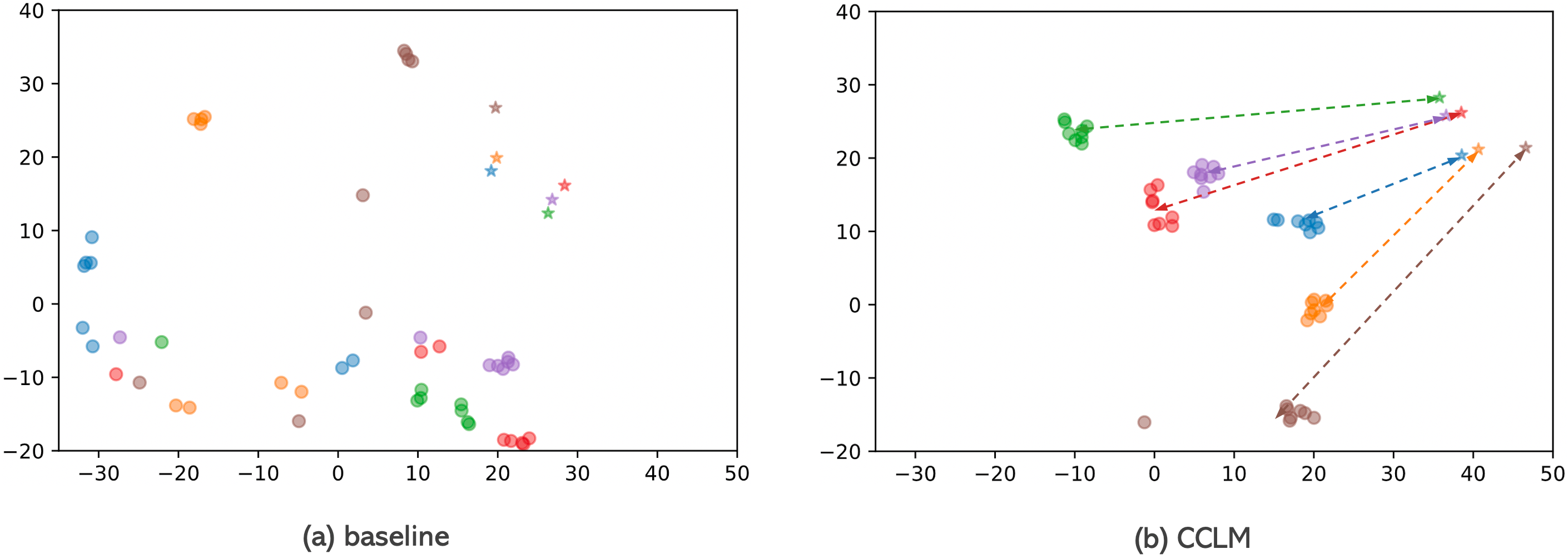}}
\caption{Visualization of image (denoted by stars) and text (denoted by points) representations. For a test example, there are eight texts in different languages. Points and stars in the same color are of the same test example. (a) is the ablated variant of \baby that does not utilize parallel sentence pairs. }
\label{Fig:features}
\end{center}
\vspace{-0.3cm}
\end{figure}

Figure \ref{Fig:features} visualizes several examples in xFlickr\&CO test set in 2D space using t-SNE~\citep{van2008visualizing}. The image representations and text representations are the output [CLS] embeddings of the image encoder and the cross-lingual text encoder respectively. 
We can observe that \baby's text representations in different languages are more gathered and the distances between text representations and corresponding image representations are relatively shorter. This indicates our approach can better align multi-lingual image-text representations.

%% file: acl2023.bbl
\begin{thebibliography}{56}
\expandafter\ifx\csname natexlab\endcsname\relax\def\natexlab#1{#1}\fi

\bibitem[{Agi{\'c} and Schluter(2018)}]{agic2017baselines}
{\v{Z}}eljko Agi{\'c} and Natalie Schluter. 2018.
\newblock \href {https://aclanthology.org/L18-1614} {Baselines and test data
  for cross-lingual inference}.
\newblock In \emph{Proceedings of the Eleventh International Conference on
  Language Resources and Evaluation ({LREC} 2018)}, Miyazaki, Japan. European
  Language Resources Association (ELRA).

\bibitem[{Bowman et~al.(2015)Bowman, Angeli, Potts, and
  Manning}]{bowman2015large}
Samuel~R. Bowman, Gabor Angeli, Christopher Potts, and Christopher~D. Manning.
  2015.
\newblock \href {https://doi.org/10.18653/v1/D15-1075} {A large annotated
  corpus for learning natural language inference}.
\newblock In \emph{Proceedings of the 2015 Conference on Empirical Methods in
  Natural Language Processing}, pages 632--642, Lisbon, Portugal. Association
  for Computational Linguistics.

\bibitem[{Brown et~al.(2020)Brown, Mann, Ryder, Subbiah, Kaplan, Dhariwal,
  Neelakantan, Shyam, Sastry, Askell, Agarwal, Herbert{-}Voss, Krueger,
  Henighan, Child, Ramesh, Ziegler, Wu, Winter, Hesse, Chen, Sigler, Litwin,
  Gray, Chess, Clark, Berner, McCandlish, Radford, Sutskever, and
  Amodei}]{brown2020language}
Tom~B. Brown, Benjamin Mann, Nick Ryder, Melanie Subbiah, Jared Kaplan,
  Prafulla Dhariwal, Arvind Neelakantan, Pranav Shyam, Girish Sastry, Amanda
  Askell, Sandhini Agarwal, Ariel Herbert{-}Voss, Gretchen Krueger, Tom
  Henighan, Rewon Child, Aditya Ramesh, Daniel~M. Ziegler, Jeffrey Wu, Clemens
  Winter, Christopher Hesse, Mark Chen, Eric Sigler, Mateusz Litwin, Scott
  Gray, Benjamin Chess, Jack Clark, Christopher Berner, Sam McCandlish, Alec
  Radford, Ilya Sutskever, and Dario Amodei. 2020.
\newblock \href
  {https://proceedings.neurips.cc/paper/2020/hash/1457c0d6bfcb4967418bfb8ac142f64a-Abstract.html}
  {Language models are few-shot learners}.
\newblock In \emph{Advances in Neural Information Processing Systems 33: Annual
  Conference on Neural Information Processing Systems 2020, NeurIPS 2020,
  December 6-12, 2020, virtual}.

\bibitem[{Bugliarello et~al.(2022)Bugliarello, Liu, Pfeiffer, Reddy, Elliott,
  Ponti, and Vuli{\'c}}]{bugliarello2022iglue}
Emanuele Bugliarello, Fangyu Liu, Jonas Pfeiffer, Siva Reddy, Desmond Elliott,
  Edoardo~Maria Ponti, and Ivan Vuli{\'c}. 2022.
\newblock \href {https://arxiv.org/abs/2201.11732} {Iglue: A benchmark for
  transfer learning across modalities, tasks, and languages}.
\newblock \emph{ArXiv preprint}, abs/2201.11732.

\bibitem[{Chen et~al.(2015)Chen, Fang, Lin, Vedantam, Gupta, Doll{\'a}r, and
  Zitnick}]{chen2015microsoft}
Xinlei Chen, Hao Fang, Tsung-Yi Lin, Ramakrishna Vedantam, Saurabh Gupta, Piotr
  Doll{\'a}r, and C~Lawrence Zitnick. 2015.
\newblock \href {https://arxiv.org/abs/1504.00325} {Microsoft coco captions:
  Data collection and evaluation server}.
\newblock \emph{ArXiv preprint}, abs/1504.00325.

\bibitem[{Chen et~al.(2020)Chen, Li, Yu, El~Kholy, Ahmed, Gan, Cheng, and
  Liu}]{chen2020uniter}
Yen-Chun Chen, Linjie Li, Licheng Yu, Ahmed El~Kholy, Faisal Ahmed, Zhe Gan,
  Yu~Cheng, and Jingjing Liu. 2020.
\newblock Uniter: Universal image-text representation learning.
\newblock In \emph{European conference on computer vision}, pages 104--120.
  Springer.

\bibitem[{Chi et~al.(2021)Chi, Dong, Wei, Yang, Singhal, Wang, Song, Mao,
  Huang, and Zhou}]{chi2020infoxlm}
Zewen Chi, Li~Dong, Furu Wei, Nan Yang, Saksham Singhal, Wenhui Wang, Xia Song,
  Xian-Ling Mao, Heyan Huang, and Ming Zhou. 2021.
\newblock \href {https://doi.org/10.18653/v1/2021.naacl-main.280} {{I}nfo{XLM}:
  An information-theoretic framework for cross-lingual language model
  pre-training}.
\newblock In \emph{Proceedings of the 2021 Conference of the North American
  Chapter of the Association for Computational Linguistics: Human Language
  Technologies}, pages 3576--3588, Online. Association for Computational
  Linguistics.

\bibitem[{Conneau et~al.(2020)Conneau, Khandelwal, Goyal, Chaudhary, Wenzek,
  Guzm{\'a}n, Grave, Ott, Zettlemoyer, and Stoyanov}]{conneau2019unsupervised}
Alexis Conneau, Kartikay Khandelwal, Naman Goyal, Vishrav Chaudhary, Guillaume
  Wenzek, Francisco Guzm{\'a}n, Edouard Grave, Myle Ott, Luke Zettlemoyer, and
  Veselin Stoyanov. 2020.
\newblock \href {https://doi.org/10.18653/v1/2020.acl-main.747} {Unsupervised
  cross-lingual representation learning at scale}.
\newblock In \emph{Proceedings of the 58th Annual Meeting of the Association
  for Computational Linguistics}, pages 8440--8451, Online. Association for
  Computational Linguistics.

\bibitem[{Conneau and Lample(2019)}]{lample2019cross}
Alexis Conneau and Guillaume Lample. 2019.
\newblock \href
  {https://proceedings.neurips.cc/paper/2019/hash/c04c19c2c2474dbf5f7ac4372c5b9af1-Abstract.html}
  {Cross-lingual language model pretraining}.
\newblock In \emph{Advances in Neural Information Processing Systems 32: Annual
  Conference on Neural Information Processing Systems 2019, NeurIPS 2019,
  December 8-14, 2019, Vancouver, BC, Canada}, pages 7057--7067.

\bibitem[{Devlin et~al.(2019)Devlin, Chang, Lee, and
  Toutanova}]{devlin2018bert}
Jacob Devlin, Ming-Wei Chang, Kenton Lee, and Kristina Toutanova. 2019.
\newblock \href {https://doi.org/10.18653/v1/N19-1423} {{BERT}: Pre-training of
  deep bidirectional transformers for language understanding}.
\newblock In \emph{Proceedings of the 2019 Conference of the North {A}merican
  Chapter of the Association for Computational Linguistics: Human Language
  Technologies, Volume 1 (Long and Short Papers)}, pages 4171--4186,
  Minneapolis, Minnesota. Association for Computational Linguistics.

\bibitem[{Dong et~al.(2019)Dong, Yang, Wang, Wei, Liu, Wang, Gao, Zhou, and
  Hon}]{DBLP:conf/nips/00040WWLWGZH19}
Li~Dong, Nan Yang, Wenhui Wang, Furu Wei, Xiaodong Liu, Yu~Wang, Jianfeng Gao,
  Ming Zhou, and Hsiao{-}Wuen Hon. 2019.
\newblock Unified language model pre-training for natural language
  understanding and generation.
\newblock In \emph{NeurIPS}, pages 13042--13054.

\bibitem[{Dosovitskiy et~al.(2021)Dosovitskiy, Beyer, Kolesnikov, Weissenborn,
  Zhai, Unterthiner, Dehghani, Minderer, Heigold, Gelly, Uszkoreit, and
  Houlsby}]{dosovitskiy2020image}
Alexey Dosovitskiy, Lucas Beyer, Alexander Kolesnikov, Dirk Weissenborn,
  Xiaohua Zhai, Thomas Unterthiner, Mostafa Dehghani, Matthias Minderer, Georg
  Heigold, Sylvain Gelly, Jakob Uszkoreit, and Neil Houlsby. 2021.
\newblock \href {https://openreview.net/forum?id=YicbFdNTTy} {An image is worth
  16x16 words: Transformers for image recognition at scale}.
\newblock In \emph{9th International Conference on Learning Representations,
  {ICLR} 2021, Virtual Event, Austria, May 3-7, 2021}. OpenReview.net.

\bibitem[{Elliott et~al.(2016)Elliott, Frank, Sima{'}an, and
  Specia}]{elliott2016multi30k}
Desmond Elliott, Stella Frank, Khalil Sima{'}an, and Lucia Specia. 2016.
\newblock \href {https://doi.org/10.18653/v1/W16-3210} {{M}ulti30{K}:
  Multilingual {E}nglish-{G}erman image descriptions}.
\newblock In \emph{Proceedings of the 5th Workshop on Vision and Language},
  pages 70--74, Berlin, Germany. Association for Computational Linguistics.

\bibitem[{Houlsby et~al.(2019)Houlsby, Giurgiu, Jastrzebski, Morrone,
  de~Laroussilhe, Gesmundo, Attariyan, and Gelly}]{houlsby2019parameter}
Neil Houlsby, Andrei Giurgiu, Stanislaw Jastrzebski, Bruna Morrone, Quentin
  de~Laroussilhe, Andrea Gesmundo, Mona Attariyan, and Sylvain Gelly. 2019.
\newblock \href {http://proceedings.mlr.press/v97/houlsby19a.html}
  {Parameter-efficient transfer learning for {NLP}}.
\newblock In \emph{Proceedings of the 36th International Conference on Machine
  Learning, {ICML} 2019, 9-15 June 2019, Long Beach, California, {USA}},
  volume~97 of \emph{Proceedings of Machine Learning Research}, pages
  2790--2799. {PMLR}.

\bibitem[{Huang et~al.(2019)Huang, Liang, Duan, Gong, Shou, Jiang, and
  Zhou}]{huang2019unicoder}
Haoyang Huang, Yaobo Liang, Nan Duan, Ming Gong, Linjun Shou, Daxin Jiang, and
  Ming Zhou. 2019.
\newblock \href {https://doi.org/10.18653/v1/D19-1252} {{U}nicoder: A universal
  language encoder by pre-training with multiple cross-lingual tasks}.
\newblock In \emph{Proceedings of the 2019 Conference on Empirical Methods in
  Natural Language Processing and the 9th International Joint Conference on
  Natural Language Processing (EMNLP-IJCNLP)}, pages 2485--2494, Hong Kong,
  China. Association for Computational Linguistics.

\bibitem[{Hudson and Manning(2019)}]{hudson2019gqa}
Drew~A. Hudson and Christopher~D. Manning. 2019.
\newblock \href {https://doi.org/10.1109/CVPR.2019.00686} {{GQA:} {A} new
  dataset for real-world visual reasoning and compositional question
  answering}.
\newblock In \emph{{IEEE} Conference on Computer Vision and Pattern
  Recognition, {CVPR} 2019, Long Beach, CA, USA, June 16-20, 2019}, pages
  6700--6709. Computer Vision Foundation / {IEEE}.

\bibitem[{Jain et~al.(2021)Jain, Guo, Srinivasan, Chen, Kudugunta, Jia, Yang,
  and Baldridge}]{jain2021mural}
Aashi Jain, Mandy Guo, Krishna Srinivasan, Ting Chen, Sneha Kudugunta, Chao
  Jia, Yinfei Yang, and Jason Baldridge. 2021.
\newblock \href {https://arxiv.org/abs/2109.05125} {Mural: multimodal,
  multitask retrieval across languages}.
\newblock \emph{ArXiv preprint}, abs/2109.05125.

\bibitem[{Karpathy and Li(2015)}]{karpathy2015deep}
Andrej Karpathy and Fei{-}Fei Li. 2015.
\newblock \href {https://doi.org/10.1109/CVPR.2015.7298932} {Deep
  visual-semantic alignments for generating image descriptions}.
\newblock In \emph{{IEEE} Conference on Computer Vision and Pattern
  Recognition, {CVPR} 2015, Boston, MA, USA, June 7-12, 2015}, pages
  3128--3137. {IEEE} Computer Society.

\bibitem[{Kong et~al.(2020)Kong, de~Masson~d'Autume, Yu, Ling, Dai, and
  Yogatama}]{KongdYLDY20}
Lingpeng Kong, Cyprien de~Masson~d'Autume, Lei Yu, Wang Ling, Zihang Dai, and
  Dani Yogatama. 2020.
\newblock \href {https://openreview.net/forum?id=Syx79eBKwr} {A mutual
  information maximization perspective of language representation learning}.
\newblock In \emph{8th International Conference on Learning Representations,
  {ICLR} 2020, Addis Ababa, Ethiopia, April 26-30, 2020}. OpenReview.net.

\bibitem[{Lewis et~al.(2020)Lewis, Liu, Goyal, Ghazvininejad, Mohamed, Levy,
  Stoyanov, and Zettlemoyer}]{lewis2019bart}
Mike Lewis, Yinhan Liu, Naman Goyal, Marjan Ghazvininejad, Abdelrahman Mohamed,
  Omer Levy, Veselin Stoyanov, and Luke Zettlemoyer. 2020.
\newblock \href {https://doi.org/10.18653/v1/2020.acl-main.703} {{BART}:
  Denoising sequence-to-sequence pre-training for natural language generation,
  translation, and comprehension}.
\newblock In \emph{Proceedings of the 58th Annual Meeting of the Association
  for Computational Linguistics}, pages 7871--7880, Online. Association for
  Computational Linguistics.

\bibitem[{Li et~al.(2021)Li, Selvaraju, Gotmare, Joty, Xiong, and
  Hoi}]{li2021align}
Junnan Li, Ramprasaath Selvaraju, Akhilesh Gotmare, Shafiq Joty, Caiming Xiong,
  and Steven Chu~Hong Hoi. 2021.
\newblock Align before fuse: Vision and language representation learning with
  momentum distillation.
\newblock \emph{Advances in Neural Information Processing Systems}, 34.

\bibitem[{Li et~al.(2019)Li, Xu, Wang, Lan, Jia, Yang, and Xu}]{li2019coco}
Xirong Li, Chaoxi Xu, Xiaoxu Wang, Weiyu Lan, Zhengxiong Jia, Gang Yang, and
  Jieping Xu. 2019.
\newblock Coco-cn for cross-lingual image tagging, captioning, and retrieval.
\newblock \emph{IEEE Transactions on Multimedia}, 21(9):2347--2360.

\bibitem[{Li et~al.(2020)Li, Yin, Li, Zhang, Hu, Zhang, Wang, Hu, Dong, Wei
  et~al.}]{li2020oscar}
Xiujun Li, Xi~Yin, Chunyuan Li, Pengchuan Zhang, Xiaowei Hu, Lei Zhang, Lijuan
  Wang, Houdong Hu, Li~Dong, Furu Wei, et~al. 2020.
\newblock Oscar: Object-semantics aligned pre-training for vision-language
  tasks.
\newblock In \emph{European Conference on Computer Vision}, pages 121--137.
  Springer.

\bibitem[{Liu et~al.(2021{\natexlab{a}})Liu, Bugliarello, Ponti, Reddy,
  Collier, and Elliott}]{liu2021visually}
Fangyu Liu, Emanuele Bugliarello, Edoardo~Maria Ponti, Siva Reddy, Nigel
  Collier, and Desmond Elliott. 2021{\natexlab{a}}.
\newblock \href {https://doi.org/10.18653/v1/2021.emnlp-main.818} {Visually
  grounded reasoning across languages and cultures}.
\newblock In \emph{Proceedings of the 2021 Conference on Empirical Methods in
  Natural Language Processing}, pages 10467--10485, Online and Punta Cana,
  Dominican Republic. Association for Computational Linguistics.

\bibitem[{Liu et~al.(2019)Liu, Ott, Goyal, Du, Joshi, Chen, Levy, Lewis,
  Zettlemoyer, and Stoyanov}]{liu2019roberta}
Yinhan Liu, Myle Ott, Naman Goyal, Jingfei Du, Mandar Joshi, Danqi Chen, Omer
  Levy, Mike Lewis, Luke Zettlemoyer, and Veselin Stoyanov. 2019.
\newblock \href {https://arxiv.org/abs/1907.11692} {Roberta: A robustly
  optimized bert pretraining approach}.
\newblock \emph{ArXiv preprint}, abs/1907.11692.

\bibitem[{Liu et~al.(2021{\natexlab{b}})Liu, Lin, Cao, Hu, Wei, Zhang, Lin, and
  Guo}]{LiuL00W0LG21}
Ze~Liu, Yutong Lin, Yue Cao, Han Hu, Yixuan Wei, Zheng Zhang, Stephen Lin, and
  Baining Guo. 2021{\natexlab{b}}.
\newblock \href {https://doi.org/10.1109/ICCV48922.2021.00986} {Swin
  transformer: Hierarchical vision transformer using shifted windows}.
\newblock In \emph{2021 {IEEE/CVF} International Conference on Computer Vision,
  {ICCV} 2021, Montreal, QC, Canada, October 10-17, 2021}, pages 9992--10002.
  {IEEE}.

\bibitem[{Loshchilov and Hutter(2019)}]{loshchilov2018decoupled}
Ilya Loshchilov and Frank Hutter. 2019.
\newblock \href {https://openreview.net/forum?id=Bkg6RiCqY7} {Decoupled weight
  decay regularization}.
\newblock In \emph{7th International Conference on Learning Representations,
  {ICLR} 2019, New Orleans, LA, USA, May 6-9, 2019}. OpenReview.net.

\bibitem[{Lu et~al.(2019)Lu, Batra, Parikh, and Lee}]{lu2019vilbert}
Jiasen Lu, Dhruv Batra, Devi Parikh, and Stefan Lee. 2019.
\newblock \href
  {https://proceedings.neurips.cc/paper/2019/hash/c74d97b01eae257e44aa9d5bade97baf-Abstract.html}
  {Vilbert: Pretraining task-agnostic visiolinguistic representations for
  vision-and-language tasks}.
\newblock In \emph{Advances in Neural Information Processing Systems 32: Annual
  Conference on Neural Information Processing Systems 2019, NeurIPS 2019,
  December 8-14, 2019, Vancouver, BC, Canada}, pages 13--23.

\bibitem[{Ni et~al.(2021)Ni, Huang, Su, Cui, Bharti, Wang, Zhang, and
  Duan}]{ni2021m3p}
Minheng Ni, Haoyang Huang, Lin Su, Edward Cui, Taroon Bharti, Lijuan Wang,
  Dongdong Zhang, and Nan Duan. 2021.
\newblock M3p: Learning universal representations via multitask multilingual
  multimodal pre-training.
\newblock In \emph{Proceedings of the IEEE/CVF Conference on Computer Vision
  and Pattern Recognition}, pages 3977--3986.

\bibitem[{Oord et~al.(2018)Oord, Li, and Vinyals}]{oord2018representation}
Aaron van~den Oord, Yazhe Li, and Oriol Vinyals. 2018.
\newblock \href {https://arxiv.org/abs/1807.03748} {Representation learning
  with contrastive predictive coding}.
\newblock \emph{ArXiv preprint}, abs/1807.03748.

\bibitem[{Peters et~al.(2018)Peters, Neumann, Iyyer, Gardner, Clark, Lee, and
  Zettlemoyer}]{peters2018elmo}
Matthew~E. Peters, Mark Neumann, Mohit Iyyer, Matt Gardner, Christopher Clark,
  Kenton Lee, and Luke Zettlemoyer. 2018.
\newblock \href {https://doi.org/10.18653/v1/N18-1202} {Deep contextualized
  word representations}.
\newblock In \emph{Proceedings of the 2018 Conference of the North {A}merican
  Chapter of the Association for Computational Linguistics: Human Language
  Technologies, Volume 1 (Long Papers)}, pages 2227--2237, New Orleans,
  Louisiana. Association for Computational Linguistics.

\bibitem[{Pfeiffer et~al.(2021)Pfeiffer, Geigle, Kamath, Steitz, Roth,
  Vuli{\'c}, and Gurevych}]{pfeiffer2021xgqa}
Jonas Pfeiffer, Gregor Geigle, Aishwarya Kamath, Jan-Martin~O Steitz, Stefan
  Roth, Ivan Vuli{\'c}, and Iryna Gurevych. 2021.
\newblock \href {https://arxiv.org/abs/2109.06082} {xgqa: Cross-lingual visual
  question answering}.
\newblock \emph{ArXiv preprint}, abs/2109.06082.

\bibitem[{Pfeiffer et~al.(2020)Pfeiffer, Vuli{\'c}, Gurevych, and
  Ruder}]{pfeiffer2020mad}
Jonas Pfeiffer, Ivan Vuli{\'c}, Iryna Gurevych, and Sebastian Ruder. 2020.
\newblock \href {https://doi.org/10.18653/v1/2020.emnlp-main.617} {{MAD-X}:
  {A}n {A}dapter-{B}ased {F}ramework for {M}ulti-{T}ask {C}ross-{L}ingual
  {T}ransfer}.
\newblock In \emph{Proceedings of the 2020 Conference on Empirical Methods in
  Natural Language Processing (EMNLP)}, pages 7654--7673, Online. Association
  for Computational Linguistics.

\bibitem[{Radford et~al.(2018)Radford, Narasimhan, Salimans, and
  Sutskever}]{radford2018improving}
Alec Radford, Karthik Narasimhan, Tim Salimans, and Ilya Sutskever. 2018.
\newblock Improving language understanding by generative pre-training.

\bibitem[{Radford et~al.(2019)Radford, Wu, Child, Luan, Amodei, Sutskever
  et~al.}]{radford2019language}
Alec Radford, Jeffrey Wu, Rewon Child, David Luan, Dario Amodei, Ilya
  Sutskever, et~al. 2019.
\newblock Language models are unsupervised multitask learners.
\newblock \emph{OpenAI blog}, 1(8):9.

\bibitem[{Raffel et~al.(2019)Raffel, Shazeer, Roberts, Lee, Narang, Matena,
  Zhou, Li, and Liu}]{raffel2019exploring}
Colin Raffel, Noam Shazeer, Adam Roberts, Katherine Lee, Sharan Narang, Michael
  Matena, Yanqi Zhou, Wei Li, and Peter~J Liu. 2019.
\newblock \href {https://arxiv.org/abs/1910.10683} {Exploring the limits of
  transfer learning with a unified text-to-text transformer}.
\newblock \emph{ArXiv preprint}, abs/1910.10683.

\bibitem[{Schwenk et~al.(2021)Schwenk, Chaudhary, Sun, Gong, and
  Guzm{\'a}n}]{schwenk2019wikimatrix}
Holger Schwenk, Vishrav Chaudhary, Shuo Sun, Hongyu Gong, and Francisco
  Guzm{\'a}n. 2021.
\newblock \href {https://doi.org/10.18653/v1/2021.eacl-main.115}
  {{W}iki{M}atrix: Mining 135{M} parallel sentences in 1620 language pairs from
  {W}ikipedia}.
\newblock In \emph{Proceedings of the 16th Conference of the European Chapter
  of the Association for Computational Linguistics: Main Volume}, pages
  1351--1361, Online. Association for Computational Linguistics.

\bibitem[{Sharma et~al.(2018)Sharma, Ding, Goodman, and
  Soricut}]{sharma2018conceptual}
Piyush Sharma, Nan Ding, Sebastian Goodman, and Radu Soricut. 2018.
\newblock \href {https://doi.org/10.18653/v1/P18-1238} {Conceptual captions: A
  cleaned, hypernymed, image alt-text dataset for automatic image captioning}.
\newblock In \emph{Proceedings of the 56th Annual Meeting of the Association
  for Computational Linguistics (Volume 1: Long Papers)}, pages 2556--2565,
  Melbourne, Australia. Association for Computational Linguistics.

\bibitem[{Srinivasan et~al.(2021)Srinivasan, Raman, Chen, Bendersky, and
  Najork}]{srinivasan2021wit}
Krishna Srinivasan, Karthik Raman, Jiecao Chen, Michael Bendersky, and Marc
  Najork. 2021.
\newblock Wit: Wikipedia-based image text dataset for multimodal multilingual
  machine learning.
\newblock In \emph{Proceedings of the 44th International ACM SIGIR Conference
  on Research and Development in Information Retrieval}, pages 2443--2449.

\bibitem[{Su et~al.(2020)Su, Zhu, Cao, Li, Lu, Wei, and Dai}]{su2019vl}
Weijie Su, Xizhou Zhu, Yue Cao, Bin Li, Lewei Lu, Furu Wei, and Jifeng Dai.
  2020.
\newblock \href {https://openreview.net/forum?id=SygXPaEYvH} {{VL-BERT:}
  pre-training of generic visual-linguistic representations}.
\newblock In \emph{8th International Conference on Learning Representations,
  {ICLR} 2020, Addis Ababa, Ethiopia, April 26-30, 2020}. OpenReview.net.

\bibitem[{Suhr et~al.(2019)Suhr, Zhou, Zhang, Zhang, Bai, and
  Artzi}]{suhr2018corpus}
Alane Suhr, Stephanie Zhou, Ally Zhang, Iris Zhang, Huajun Bai, and Yoav Artzi.
  2019.
\newblock \href {https://doi.org/10.18653/v1/P19-1644} {A corpus for reasoning
  about natural language grounded in photographs}.
\newblock In \emph{Proceedings of the 57th Annual Meeting of the Association
  for Computational Linguistics}, pages 6418--6428, Florence, Italy.
  Association for Computational Linguistics.

\bibitem[{Tan and Bansal(2019)}]{tan2019lxmert}
Hao Tan and Mohit Bansal. 2019.
\newblock \href {https://doi.org/10.18653/v1/D19-1514} {{LXMERT}: Learning
  cross-modality encoder representations from transformers}.
\newblock In \emph{Proceedings of the 2019 Conference on Empirical Methods in
  Natural Language Processing and the 9th International Joint Conference on
  Natural Language Processing (EMNLP-IJCNLP)}, pages 5100--5111, Hong Kong,
  China. Association for Computational Linguistics.

\bibitem[{Tiedemann(2012)}]{tiedemann2012parallel}
J{\"o}rg Tiedemann. 2012.
\newblock Parallel data, tools and interfaces in opus.
\newblock In \emph{Lrec}, volume 2012, pages 2214--2218. Citeseer.

\bibitem[{Van~der Maaten and Hinton(2008)}]{van2008visualizing}
Laurens Van~der Maaten and Geoffrey Hinton. 2008.
\newblock Visualizing data using t-sne.
\newblock \emph{Journal of machine learning research}, 9(11).

\bibitem[{Wang et~al.(2022)Wang, Yang, Men, Lin, Bai, Li, Ma, Zhou, Zhou, and
  Yang}]{wang2022ofa}
Peng Wang, An~Yang, Rui Men, Junyang Lin, Shuai Bai, Zhikang Li, Jianxin Ma,
  Chang Zhou, Jingren Zhou, and Hongxia Yang. 2022.
\newblock Ofa: Unifying architectures, tasks, and modalities through a simple
  sequence-to-sequence learning framework.
\newblock In \emph{International Conference on Machine Learning}, pages
  23318--23340. PMLR.

\bibitem[{Xie et~al.(2019)Xie, Lai, Doran, and Kadav}]{xie2019visual}
Ning Xie, Farley Lai, Derek Doran, and Asim Kadav. 2019.
\newblock \href {https://arxiv.org/abs/1901.06706} {Visual entailment: A novel
  task for fine-grained image understanding}.
\newblock \emph{ArXiv preprint}, abs/1901.06706.

\bibitem[{Yoshikawa et~al.(2017)Yoshikawa, Shigeto, and
  Takeuchi}]{yoshikawa2017stair}
Yuya Yoshikawa, Yutaro Shigeto, and Akikazu Takeuchi. 2017.
\newblock \href {https://doi.org/10.18653/v1/P17-2066} {{STAIR} captions:
  Constructing a large-scale {J}apanese image caption dataset}.
\newblock In \emph{Proceedings of the 55th Annual Meeting of the Association
  for Computational Linguistics (Volume 2: Short Papers)}, pages 417--421,
  Vancouver, Canada. Association for Computational Linguistics.

\bibitem[{Young et~al.(2014)Young, Lai, Hodosh, and
  Hockenmaier}]{young2014image}
Peter Young, Alice Lai, Micah Hodosh, and Julia Hockenmaier. 2014.
\newblock \href {https://doi.org/10.1162/tacl_a_00166} {From image descriptions
  to visual denotations: New similarity metrics for semantic inference over
  event descriptions}.
\newblock \emph{Transactions of the Association for Computational Linguistics},
  2:67--78.

\bibitem[{Yu et~al.(2022)Yu, Wang, Vasudevan, Yeung, Seyedhosseini, and
  Wu}]{yu2022coca}
Jiahui Yu, Zirui Wang, Vijay Vasudevan, Legg Yeung, Mojtaba Seyedhosseini, and
  Yonghui Wu. 2022.
\newblock Coca: Contrastive captioners are image-text foundation models.
\newblock \emph{arXiv preprint arXiv:2205.01917}.

\bibitem[{Zeng(2021)}]{zeng-2021-multi}
Danting Zeng. 2021.
\newblock \href {https://doi.org/10.18653/v1/2021.maiworkshop-1.5} {Multi task
  learning based framework for multimodal classification}.
\newblock In \emph{Proceedings of the Third Workshop on Multimodal Artificial
  Intelligence}, pages 30--35, Mexico City, Mexico. Association for
  Computational Linguistics.

\bibitem[{Zeng and Nie(2021)}]{zeng-nie-2021-investigation}
Yan Zeng and Jian-Yun Nie. 2021.
\newblock \href {https://doi.org/10.18653/v1/2021.findings-acl.393} {An
  investigation of suitability of pre-trained language models for dialogue
  generation {--} avoiding discrepancies}.
\newblock In \emph{Findings of the Association for Computational Linguistics:
  ACL-IJCNLP 2021}, pages 4481--4494, Online. Association for Computational
  Linguistics.

\bibitem[{Zeng et~al.(2021)Zeng, Zhang, and Li}]{zeng2021multi}
Yan Zeng, Xinsong Zhang, and Hang Li. 2021.
\newblock \href {https://arxiv.org/abs/2111.08276} {Multi-grained vision
  language pre-training: Aligning texts with visual concepts}.
\newblock \emph{ArXiv preprint}, abs/2111.08276.

\bibitem[{Zeng et~al.(2022)Zeng, Zhang, Li, Wang, Zhang, and Zhou}]{zeng2022x}
Yan Zeng, Xinsong Zhang, Hang Li, Jiawei Wang, Jipeng Zhang, and Wangchunshu
  Zhou. 2022.
\newblock X$^2$-vlm: All-in-one pre-trained model for vision-language tasks.
\newblock \emph{arXiv preprint arXiv:2211.12402}.

\bibitem[{Zhou et~al.(2021)Zhou, Zhou, Wang, Cheng, Li, Yu, and
  Liu}]{zhou2021uc2}
Mingyang Zhou, Luowei Zhou, Shuohang Wang, Yu~Cheng, Linjie Li, Zhou Yu, and
  Jingjing Liu. 2021.
\newblock Uc2: Universal cross-lingual cross-modal vision-and-language
  pre-training.
\newblock In \emph{Proceedings of the IEEE/CVF Conference on Computer Vision
  and Pattern Recognition}, pages 4155--4165.

\bibitem[{Zhou et~al.(2022)Zhou, Zeng, Diao, and Zhang}]{zhou2022vlue}
Wangchunshu Zhou, Yan Zeng, Shizhe Diao, and Xinsong Zhang. 2022.
\newblock \href {http://arxiv.org/abs/2205.15237} {Vlue: A multi-task benchmark
  for evaluating vision-language models}.
\newblock \emph{CoRR}, abs/2205.15237.

\bibitem[{Ziemski et~al.(2016)Ziemski, Junczys-Dowmunt, and
  Pouliquen}]{ziemski2016united}
Micha{\l} Ziemski, Marcin Junczys-Dowmunt, and Bruno Pouliquen. 2016.
\newblock The united nations parallel corpus v1. 0.
\newblock In \emph{Proceedings of the Tenth International Conference on
  Language Resources and Evaluation (LREC'16)}, pages 3530--3534.

\end{thebibliography}
